\pdfoutput=1

\documentclass[11pt]{article}

\usepackage[preprint]{acl}

\usepackage{times}
\usepackage{latexsym}

\usepackage[T1]{fontenc}

\usepackage[utf8]{inputenc}

\usepackage{microtype}

\usepackage{inconsolata}

\usepackage{graphicx}

\usepackage{adjustbox}
\usepackage{booktabs}
\usepackage{xcolor}
\usepackage{enumitem}
\usepackage{longtable}
\usepackage{multirow}
\usepackage{amsmath}
\usepackage{multicol}
\usepackage{tabularx}
\usepackage[colorinlistoftodos]{todonotes}

\title{A Survey on LLM-Assisted Clinical Trial Recruitment}

\author{Shrestha Ghosh$^{1}$
\qquad
Moritz Schneider$^{2}$
\qquad
Carina Reinicke$^{2}$
\qquad
Carsten Eickhoff$^{1}$
\\
\\
$^{1}$University of Tübingen, Germany
\\
$^{2}$Boehringer Ingelheim, Germany
\\
{\small$^{1}$\texttt{\{first.last\}@uni-tuebingen.de}
\qquad $^{2}$\texttt{\{first.last\}@boehringer-ingelheim.com}}
}

\begin{document}

\newcommand{\sg}[2]{{\color{blue}#1: #2}}
\renewcommand{\paragraph}[1]{\smallskip\par\noindent\textbf{#1}.}

\maketitle
\begin{abstract} 
  Clinical trials are designed in natural language and the task of matching them to patients, represented via both structured and unstructured textual data, benefits from knowledge aggregation and reasoning abilities of LLMs. 
  LLMs with their ability to consolidate distributed knowledge hold the potential to build a more general solution than classical approaches that employ trial-specific heuristics. 
  Yet, adoption of LLMs in critical domains, such as clinical research, comes with many challenges, such as, the size of the benchmarks, the dimensions of evaluation and data sensitivity.
  In this survey, we contextualize emerging LLM-based approaches in clinical trial recruitment. We examine the main components of the clinical trial recruitment process, discuss existing challenges in adopting LLM technologies in clinical research and exciting future directions.
\end{abstract}

\section{Introduction}
Clinical trials evaluate the effects of an intervention on human health. Selecting the precise and required size of patient population is crucial for trial completion.  
According to various estimates, more than 50\% of aborted clinical trials fail due to low accrual rates, and 80\% of all clinical trials do not manage to recruit the required patient cohorts within the allotted time \cite{clinicaltrialarena2012,williams2015terminated,pharmaceutical2019}. While the trend has steadily declined over the past decade, with the intensive use of technology-aided solutions, efficient patient recruitment remains the most crucial bottleneck in clinical trial research \cite{clinicaltrialarena2022}.
As electronic health records (EHRs) of patients become more accessible, clinical researchers adopt machine intelligence and develop explainable systems to correctly interpret model predictions \cite{murdoch2013inevitable,payrovnaziri2020explainable, vonitzstein2021application}.

There are methods leveraging LLMs in clinical trial research for cohort retrieval and modelling \cite{fang2022combining, tian2023parsing,park2024criteria2query,liu2025neuralcohort,wang2025trialpanorama}, trial design \cite{reinisch2024ctp,curran2024examining,bornet2025analysis,neehal2025large}, trial search \cite{white2023clinidigest,rybinski2020clinical}, trial matching \cite{jin2024matching,nievas2024distilling,wornow2025zero}, predicting trial outcomes and duration \cite{reinisch2024ctp,yue2024clinicalagent,yue2024trialdura,liu2025autoct}, assessing risk of bias \cite{lai2024assessing,ji2025robguard}, and extracting clinical trial results \cite{lee2024seetrials}, even as the community is coming up with recommended practices for responsible use of AI in the entire drug development process \cite{geraci2025current}.

\begin{figure*}[t]
    \centering
    \includegraphics[width=0.95\linewidth, trim=0 3cm 1cm 0]{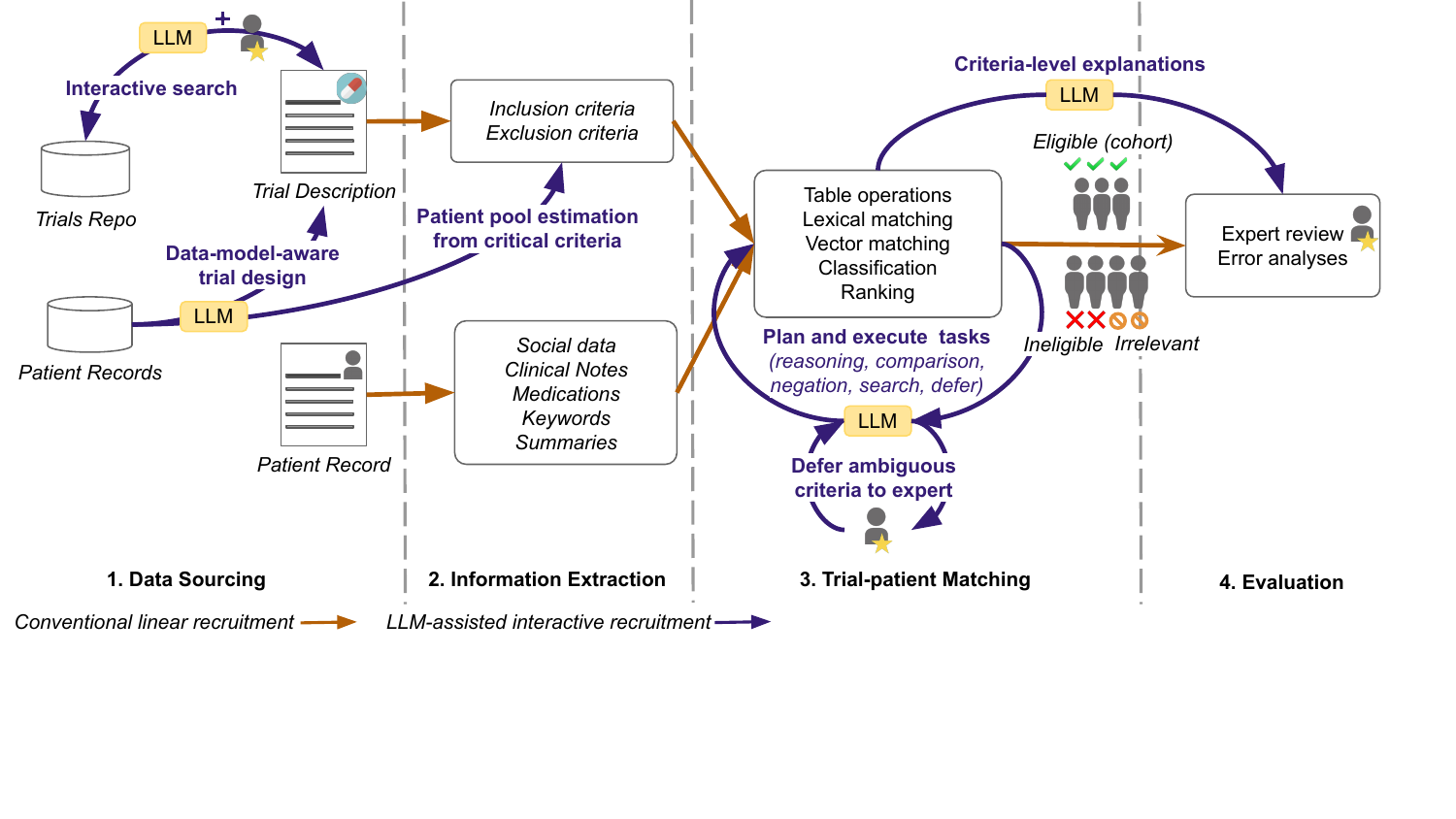}
    \caption{Components in a patient recruitment process: conventional linear flow (in orange) vs. our proposed LLM-assisted interactive flow (in purple).}
    \label{fig:patient-recruitment-proposed}
\end{figure*}

Figure~\ref{fig:patient-recruitment-proposed} shows the components in clinical trial recruitment, namely, data sourcing, information extraction, matching, and evaluation.
An expert reviews several hundred patients per trial and can end up spending hours on one patient, hence incurring significant costs \cite{penberthy2012effort,ni2015increasing}. Even simple automation using table queries and lexical searches saves between 165 hours to 1,329 hours of reviewing time when compared to manual evaluation \citet{penberthy2010automated}.
In the past, the patient recruitment process has seen relatively low adoption of the pretrained language models \cite{he2020clinical, harrer2023attention, lu2024large}. 
Generative LLMs serve as knowledge aggregators, and
through their reasoning and instruction-following capabilities, they have revived research in the task of trial and patient matching \cite{jin2024matching, nievas2024distilling,rybinski2024learning,wornow2025zero}. 

\paragraph{Difference to Prior Work} Despite the rapidly evolving landscape of LLM technology, there is no prior work surveying this area. 
\citet{gueguen2025prospective} evaluate public trial matching tools and \citet{layne2025large} compare the efficacy of open and proprietary LLM-assisted trial-patient matching in oncology.
Systematic reviews in this topic, bound by strict selection criteria and highly specific research question, do not to capture the broad perspectives \cite{kim2022review,idnay2022systematic,kantor2024machine,chen2025enhancing} ({Appendix \ref{appen:survey_guidelines} differentiates between systematic reviews and surveys}).
Although it provides an overview of LLM approaches used in clinical trial matching, \citet{chen2025enhancing} only briefly discusses associated challenges and future work. 

%

\paragraph{Our Contribution}
We examine the main components in a trial recruitment process, presented in Figure \ref{fig:patient-recruitment-proposed}. We formalize the problem of trial-patient matching. We analyze existing approaches via the tasks (classification vs. ranking), the directionality (trial-centric and patient-centric) and the choice of benchmarks (longitudinal vs. short patient descriptions, single vs. multi-trial). We present a taxonomy of errors for consistent evaluation of LLM-generated responses. We discuss the critical challenges associated with using LLMs in trial recruitment research. Finally, we present actionable steps towards an interactive patient recruitment (illustrated in purple in Figure \ref{fig:patient-recruitment-proposed}).

\section{Background}\label{sec:background}
\paragraph{Clinical Trial Recruitment} Also known as patient recruitment/enrollment/(pre-)screening, it is the process of matching patients (or a cohort) to a clinical trial via its eligibility criteria. 
Clinical trials have a dual nature, consisting of both universal and trial-specific requirements, making it challenging to design generalized approaches \cite{idnay2023clinical}. 
This has traditionally resulted in linear trial-centric matching processes (illustrated in orange in Figure \ref{fig:patient-recruitment-proposed}) with limited scope for interaction and feedback. Such systems generally have an initial data filter on the structured EHRs followed by keyword matches and concept identification \cite{penberthy2010automated,tun2023automatic} or cohort-specific classifiers \cite{zhang2017automated}. \citet{ni2015increasing} represented both trials and patients as feature vectors, thus supporting retrievals for both trial and patient.

\paragraph{Biomedical NLP and LLMs}\label{sec:bg-par:biomedicalnlp}
The biomedical NLP landscape is shifting from specialized pretrained language models, such as,
BioBERT \cite{lee2020biobert}, BioLM \cite{lewis2020pretrained}, PubmedBERT \cite{gu2021domain}, BioGPT \cite{luo2022biogpt}, MedCPT \cite{jin2023medcpt}, among many others \cite{wang2023pre}, towards instruction-following and chat-enabled LLMs (commonly termed as generative AI), used as is \cite{nori2023capabilities,kung2023performance} or fine-tuned for domain alignment, such as, Med-PaLM \cite{singhal2023large}, Med-Alpaca \cite{han2023medalpaca} and LLaVA-Med \cite{li2023llava}. We point our reader to \citet{thirunavukarasu2023large, liu2025application} for further reading.
The Journal of American Medical Informatics Association (JAMIA), which published 41 articles on biomedical health and LLMs in a focus issue, also observed this shift towards generative AI \cite{lu2024large}. Among these, the OpenAI \cite{achiam2023gpt} family of proprietary models (GPT3.5 and GPT4 being the most common) is used far more often than open-sourced models \cite{touvron2023llama,bai2023qwen,jiang2023mistral} models. 
Further, fine-tuning generative LLMs on the biomedical domain offers limited performance gain \citet{dorfner2024biomedical}. 
In fact, \citet{alber2025medical} show that models are more prone to propagate medical misinformation encountered during fine-tuning despite performing well on benchmarks.
Finally, \citet{bedi2025testing} report that real patient data is used in less than 5\% of the 519 biomedical studies using LLMs, and fairness, bias, uncertainty and deployment considerations are rarely assessed.

\paragraph{State of Adoption of NLP Advancements} 
\citet{kantor2024machine}'s study on AI adoption rates in eligibility criteria parsing, report very low adoption rates of generative AI, with BERT family of models being most popular and generative models coming into the picture only since 2024.
 A systematic review of the role of NLP systems in patient recruitment identified only 11 studies \cite{idnay2022systematic}. Heterogeneous outcomes, diverse results, a dependence on small retrospective data and a lack of common standardized benchmarks drive the gap in NLP research and their adoption in real-world settings \cite{idnay2024clinical, kantor2024machine}. 
A study by \citet{idnay2023clinical}, investigating how clinical researchers screened patients, highlights the challenges of universal and domain-specific nature of the eligibility criteria and makes recommendations to build interactive, flexible and transparent recruitment strategies. Interestingly, when \citet{corbaux2024patients} categorize tools for oncological trial matching, they indicate that the automatic methods still fall in the research and development phase, yet to be commercially available.

\section{Methodology}
We queried Google Scholar with the keywords “clinical trial”, “cohort discovery”, “patient recruitment”, “trial recruitment”, “trial matching” in conjunction with “llm”, “language model”, “gpt” and inspected the first ten result pages, between the years 2019 and 2025 (both inclusive). We also used \citet{jin2024matching} and \citet{wornow2025zero} as seeds and recursively traced their citations to efficiently captured the evolution of the predictive methods in this field.
We organized papers that described LLM-based approaches and corpora for clinical trial recruitment into data sourcing (Section \ref{sec:background}), information extraction and parsing (Section \ref{sec:information-extraction}), trial-patient matching (Section \ref{sec:trial-patient-match}) and evaluation (Section \ref{sec:evaluation}). 
Via our discussion on critical limitations in Section \ref{sec:discussion} and on promising directions towards an interactive patient recruitment in Section \ref{sec:future_directions}, we provide further contextualize LLMs in healthcare settings and other stages of the drug development process that directly effect on patient recruitment.

\begin{table*}[t]
    
    \begin{adjustbox}{width=0.9\linewidth, center}
    \small
    \begin{tabular}{@{}p{1cm} p{2.1cm} p{5cm} p{2.3cm} @{}p{2.1cm} p{0.6cm}@{} p{3cm}@{}}
    \toprule
       \textbf{Task} &  \textbf{Benchmark} & \textbf{Size} & \textbf{Eval. Metrics} & \textbf{Best Score} & \textbf{\#P} & \textbf{LLM Usage (\#P)} \\
       \midrule
        
        Cohort\newline selection & 2018 N2C2 \newline\cite{stubbs2019cohort}  & 
        \#patients = 288 
        \newline\#records per patient = 2-5 
        \newline\#tokens/patient = 2711
        \newline \#trials = 1 
        \newline\#criteria = 13&
        Micro-averaged:
        \newline P, R, F1 &
        0.91 micro F1 &
        47 &
        \textit{(not applicable)}
        \\
        \midrule
        
        \multirow{4}{2cm}{Trial\newline ranking} & \citet{koopman2016test}  & 
        \#patient summaries (22 words avg) = 60
        \newline\#patient description (78 words avg) = 60
        \newline\#keywords (4.4 words avg) = 489
        \newline\#trials = 204,855 
        \newline\#relevant matches = 685
        \newline\#relevance judgments = 4000
        & MRR,
        \newline P@5, 
        \newline adaptive precision & 
        0.3 MRR
        \newline$<$ 0.2 P@5 &
        - $\dagger$ & \textit{(not applicable)} \\
        \cmidrule{3-7}
        
        & TREC CT 2021\newline\cite{soboroff2021overview}$^*$ & 
        \#patient descriptions = 75
        \newline\#trials = 375,580 
        \newline\#relevant matches = 5,570
        \newline\#relevance judgments = 35,832&
        NDCG@10,
        \newline P@10,  
        \newline MRR, 
        \newline R-Precision
        & 0.84 NDCG@10 
        \newline 0.748 P@10
        \newline 1.0 MRR &
        26 &
        BERT-based keyword extraction (9) / query summarization (1),
        \newline Transformer-based rankers (10)\\
        \cmidrule{3-7}

        & TREC CT 2022\newline\cite{roberts2022overview}  &
        \#patient descriptions = 50 
        \newline trials = 375,580
        \newline\#relevant matches = 3,949
        \newline\#relevance judgments = 35,394&
        NDCG@10,
        \newline P@10,
        \newline MRR,
        \newline R-Precision&
        0.61 NDCG@10
        \newline 0.50 P@10
        \newline 0.32 R-Precision
        \newline 0.72 MRR &
        12 &
        Query reformulation using BERT / sequence-to-sequence models (3),
        \newline Transformer-based rankers (3)\\
        \cmidrule{3-7}
        
        & TREC CT 2023\newline\cite{rybinski2024learning}$^{**}$  &
        \#patient tables = 40
        \newline\#disease templates = 8
        \newline\#trials = 451,538
        \newline\#relevant matches = 11,963
        \newline\#relevance judgments = 34,931&
        NDCG@10,
        \newline P@10,
        \newline MRR & 0.81 NDCG@10
        \newline 0.73 P@10
        \newline 0.78 MRR &
        11 &
        Query reformulation using LLMs (5),
        \newline Transformer-based rankers (6),
        \newline LLM prompt-based relevance prediction (4)\\
        \bottomrule
        \multicolumn{7}{l}{\textit{$^*$Best scores are aggregated from the Appendix of the proceedings of TREC 2021.}
        \hfill\textit{$^\dagger$No participants, since this was not a challenge.}}\\
        \multicolumn{7}{l}{\textit{$^{**}$Borrowed from \cite{rybinski2024learning} as TREC CT 2023 does not have a published track overview.}}
    \end{tabular}
    \end{adjustbox}
    \caption{Overview of the public trial-patient matching benchmarks. \#P is the number of participating teams. LLM usage (\#P) tracks the number of participants using LLMs that we could verify from the proceedings.} 
    \label{tab:benchmarks_and_evaluation}
\end{table*}

\section{Data Sourcing: Public Benchmarks}\label{sec:benchmarks}
We analyze five trial-patient matching benchmarks.
\begin{itemize}[noitemsep,labelindent=0pt,leftmargin=1em,topsep=0pt]
    \item 2018 N2C2 Cohort Selection \cite{stubbs2019cohort} for criterion-level eligibility prediction.\footnote{Currently unavailable as of 2024 Nov 6. 
    }
    \item \citet{koopman2016test} for ranking trials.
    \item Text REtrieval Conference Clinical Trial (TREC CT)\footnote{Available at \url{https://www.trec-cds.org/}.} tracks 2021, 2022, 2023 for ranking trials.
\end{itemize} 
 Table \ref{tab:benchmarks_and_evaluation} provides an overview of the benchmarks.

\subsection{Trial Data}
The TREC CT benchmarks and \citet{koopman2016test} source the \url{ClinicalTrials.gov} registry. Meanwhile, the N2C2 benchmark focuses on a single trial. 
\url{ClinicalTrials.gov} is one of the largest online databases of clinical studies submitted by investigators from over 200 countries. {As of June 2025, it lists more than 400,000 clinical trials, thousands of which are active}. The \url{euclinicaltrials.eu} is another online registry comprising over 50,000 trials from the European Union of which, 7000 are active. Every trial comprises a study description, eligibility criteria and study plan among other details.

\subsection{Patient Data} 
The N2C2 benchmark contains 288 de-identified longitudinal records of patients. The trial ranking benchmarks use up to 75 synthetic patient profiles, comprising keywords defining cohorts \citet{koopman2016test}, short textual patient descriptions (TREC CT 2021, 2022) and questionnaire templates (TREC CT 2023). 
The MIMIC database \cite{johnson2016mimic, johnson2020mimic} is the largest anonymized public database of structured and unstructured data of 299,712 patients.
While this dataset is popular for training biomedical LLMs (as mentioned in Section \ref{sec:background}),
it is not used in any of the five benchmarks, possibly due to significant challenges in obtaining eligibility labels at this scale.

\subsection{Annotated Labels}
Two medical experts annotated 3,744 criterion-level labels in the N2C2 benchmark.
The TREC CT annotations were created by pooling the top-k results from all participating teams. 
Medical experts manually annotated this pool of trial-patient matches  \cite{koopman2016test,soboroff2021overview, roberts2022overview}.
Out of a total of 35,394 pooled trial-patient matches in the 2022 edition, 11\% were judged as \textit{eligible}, 9\% as \textit{ineligible} and 80\% as \textit{not relevant} with an average of 700 trials judged per patient \cite{roberts2022overview}. The relevance judgments were more balanced between the three labels \cite{rybinski2024learning} in TREC CT 2023. 
All benchmarks depend on manual annotation from experts, which is time-consuming and challenging to scale \cite{kim2022review}.

\section{Information Extraction and Parsing}\label{sec:information-extraction}
Extraction of medical entities evolved from a combination of rule-based heuristics and feature-based supervised sequence labelling models \cite{kang2017eliie,yuan2019criteria2query} via embedding-based neural models \cite{khan2019improving,tseo2020information} to transformer-based pretrained models and LLMs \cite{liu2021clinical,zeng2020ensemble,tian2021transformer,li2022comparative,murcia2024automating,kantor2024finetuned}. \citet{datta2024autocriteria} use disease-specific prompting to extract structured information about entities and its attributes from criteria text. \citet{gao2020compose,zhang2020deepenroll,theodorou2023treement} use BERT-based embeddings to encode eligibility criteria and patient data.
Patient data converted to search queries, via reformulation and expansion using LLMs, are particularly effective in trial retrieval \cite{peikos2023investigating,rybinski2024learning,peikos2024leveraging,jin2024matching}. \citet{yuan2019criteria2query,tian2023parsing,park2024criteria2query,mugambi2024leveraging,ziletti2025generating} use semantic parsing to translate eligibility criteria into logical forms ready for querying patient databases.

\section{Trial-Patient Matching}\label{sec:trial-patient-match}
\subsection{Formalization}
Given sets of inclusion and exclusion criteria ($C_{inc}, C_{exc}$) from a trial and a set of patient data, $P$, we formalize the trial-patient matching problem, $M$, to predict one of the labels \textit{Ineligible (Inel.)}, \textit{Irrelevant (Irr.)} or \textit{Eligible (Eli.)} by aggregating criterion-level binary matches $M'(c,P)$.
\!
\[
\small
\!M(C_{inc}, C_{exc},P)\!=\!\left\{\!\begin{array}{ll}
        \!\textit{Inel.},\!&\!\exists c \in C_{exc}, M'(c,P)\\
        \!\textit{Irr.},\!&\!\exists c \in C_{inc}, \neg M'(c,P)\\
        \!\textit{Eli.},\!&\!\neg(\textit{Inel.} \lor \textit{Irr.})
        \end{array}\!\right.\label{eq:matching}\tag{1}
\]

This induces a priority, such that, a patient satisfying any exclusion criteria becomes ineligible, regardless of inclusion criteria matches. 
Section \ref{subsec:llm_approaches} explores the task at varying levels of granularity, starting from criterion-level via trial-level predictions to trial ranking.
In Section \ref{subsec:discussion_formalization}, we elaborate on the importance of formalization and its effects on trial-level aggregation.

Despite the task being direction-agnostic, there exist two directional approaches to tackle the matching problem, mainly due to data availability.
First is the \textbf{trial-centric} approach, taken by a trial investigator, that matches longitudinal patient records to a specific trial. The 2018 N2C2 cohort selection is a trial-centric benchmark with criterion-level predictions.
Alternatively, a \textbf{patient-centric} approach, taken by a patient or their healthcare provider, matches relevant trials from a trial registry to a short description of the patient. \citet{koopman2016test} and the TREC CT 2021, 2022, and 2023 are patient-centric benchmarks for ranking trials.

\begin{table}[t]
    \centering
    \begin{adjustbox}{max width=0.9\linewidth}
    \renewcommand{\arraystretch}{0.7}
    \begin{tabularx}{\columnwidth}{@{}XX@{}}
    \toprule
    \textbf{Classical} & \textbf{LLM-based} \\
    \midrule
    {\color{red}\textbf{!}} Direction-specific approaches, applicable to a set of trials or a cohort & {\color{teal}\textbf{*}} Direction-agnostic criteria- \& trial-level prediction\\
    {\color{red}\textbf{!}} Trial-specific heuristics: filters and features & {\color{teal}\textbf{*}} Generalizable across trials\\
    {\color{red}\textbf{!}} Evaluated on private patient data & {\color{teal}\textbf{*}} Public benchmarks more common \\ 
    \bottomrule
    \end{tabularx}
    \end{adjustbox}
    \caption{Differences between classical and LLM-based trial matching approaches.}
    \label{tab:compare_classical_vs_llm_based}
\end{table}

\subsection{LLM-based Approaches}\label{subsec:llm_approaches} 
Unlike classical approaches \cite{penberthy2010automated, zhang2017automated, yuan2019criteria2query, tun2023automatic}, LLM-based approaches do not rely on trial-specific heuristics. Table \ref{tab:compare_classical_vs_llm_based} lists the primary differences between the classical and LLM-based approaches (see Appendix \ref{appendix:table_matching_systems} for a full comparison). We group the LLM-based approaches by the granularity of the matches, starting with criterion-level prediction via trial-level prediction to trial ranking. While criterion- and trial-level predictions are direction-agnostic, trial ranking is a patient-centric task. With enough patients, we could evaluate patient ranking, similar to the bidirectional implementation in \citet{ni2015increasing}'s work, though this has not yet been addressed.


\paragraph{Criterion-Level Prediction}
Here, methods utilize the reasoning capability of LLMs to obtain decision rationale and other context data in addition to the eligibility label.
\citet{hamer2023improving} use 1-shot prompt, where given the patient profile and the eligibility criteria, the LLM first labels each criterion as being applicable to the patient or not, followed by a list of rationales and finally, the eligibility labels for the applicable criteria. 
\citet{unlu2024retrieval,beattie2024utilizing} and 
\citet{wornow2025zero} chunk longitudinal patient data and store them as vector embeddings. They prompt LLMs with criteria and relevant patient data chunks under zero-shot setting, with \citet{beattie2024utilizing} providing expert criterion-level strategy in the prompts. \citet{unlu2024retrieval} generate only the decision label from GPT4 and \citet{beattie2024utilizing} and \cite{wornow2025zero} generate criterion-level JSON objects, comprising the criteria label, the eligibility label and a rationale among other context. Both work with proprietary OpenAI models (GPT-3.5 and GPT-4) and \citet{wornow2025zero} additionally used open-sourced Llama \cite{jiang2023mistral} and Mixtral models \cite{jiang2024mixtral}. \citet{ferber2024end} prompt GPT-4o to generate criterion-level boolean predictions to be reviewed by experts.  

\paragraph{Trial-Level Prediction}
\citet{wong2023scaling} used GPT3.5 and GPT4 to extract and structure eligibility criteria into logical expression to be matched locally with structured patient information.
\citet{yuan2023large} use LLMs to augment eligibility criteria and pass the BERT-based embeddings of these criteria and patient data through a fully connected classification layer to predict patient-criterion eligibility. The trial-level prediction is then computed by asserting Equation \ref{eq:matching}.  

\paragraph{Trial Ranking}
Here, methods formulate effective queries, retrieve trials and re-rank them.
Query processing techniques involve generating sentence queries from patient descriptions using a fine-tuned T5 model \cite{pradeep2022neural} or zero-shot LLM prompts \cite{saeidi2023malnis,kusa2023dossier}, generating patient descriptions from trial data via 1-shot LLM prompts \cite{zhuang2023team}, generating no-SQL queries via LLMs \cite{ferber2024end}, using LLMs to reformulate and expand queries \cite{rybinski2024learning,peikos2024leveraging,datta2025patient2trial}, and using LLMs to extract keywords \cite{jullien2024controlled,jin2024matching,nievas2024distilling}.
This is followed by retrieval, either using embeddings similarity only \cite{lahiri2023tmu,richmond2023leveraging,saeidi2023malnis,ferber2024end,saeidi2025streamlining} or a multi-stage retrieval with neural re-rankers \cite{zhuang2023team,rybinski2023matching,rybinski2024learning,jin2024matching,datta2025patient2trial}. Certain approaches re-rank the top trials using GPT models \cite{zhuang2023team,rybinski2024learning,datta2025patient2trial}, some prompt LLMs for relevance labels \cite{pradeep2022neural}, while others prompt LLMs to generate trial- or criterion-level eligibility labels \cite{rybinski2024learning,piekos2023unimib,jin2024matching,nievas2024distilling,jullien2024controlled}. Criterion-level labels are aggregated using set-based reasoning mechanisms \cite{jullien2024controlled}, variations of the matching Equation \ref{eq:matching} \cite{jin2024matching,nievas2024distilling,saeidi2025streamlining}, and by prompting LLMs \cite{jin2024matching}.

\section{Evaluation}\label{sec:evaluation}
Standard IR metrics, such as, precision, recall, F1, precision at rank k, normalized discounted cumulative gains (NDCG@k) and mean reciprocal rank (MRR), are the most popular evaluation metrics. 
The 2018 N2C2 cohort selection and TREC CT benchmarks have high performance metrics (Table \ref{tab:benchmarks_and_evaluation}), which are further improved by newer methods. 
Notably, researchers in the medical domain have stressed the lack of consistent benchmark data, primarily, due to the dependence on manual review which also restricts the size of the data \cite{kim2022review,kantor2024machine}.

Besides accuracy metrics, \citet{rybinski2024learning} report query latency, with GPT models having 15 times higher latency (47.8s) when compared to smaller pre-trained models for a 12\% increase in NDCG@10. \citet{jin2024matching} report a 42\% reduction in screening time when using LLM-generated predictions and explanations.
The measurement of human preference of LLM generations, bias in model predictions and the informativeness of LLM explanations ranges from limited to non-existent. 

\begin{figure}[t]
    \centering
    \includegraphics[width=0.9\linewidth, trim=0 3.5cm 13.6cm 0]{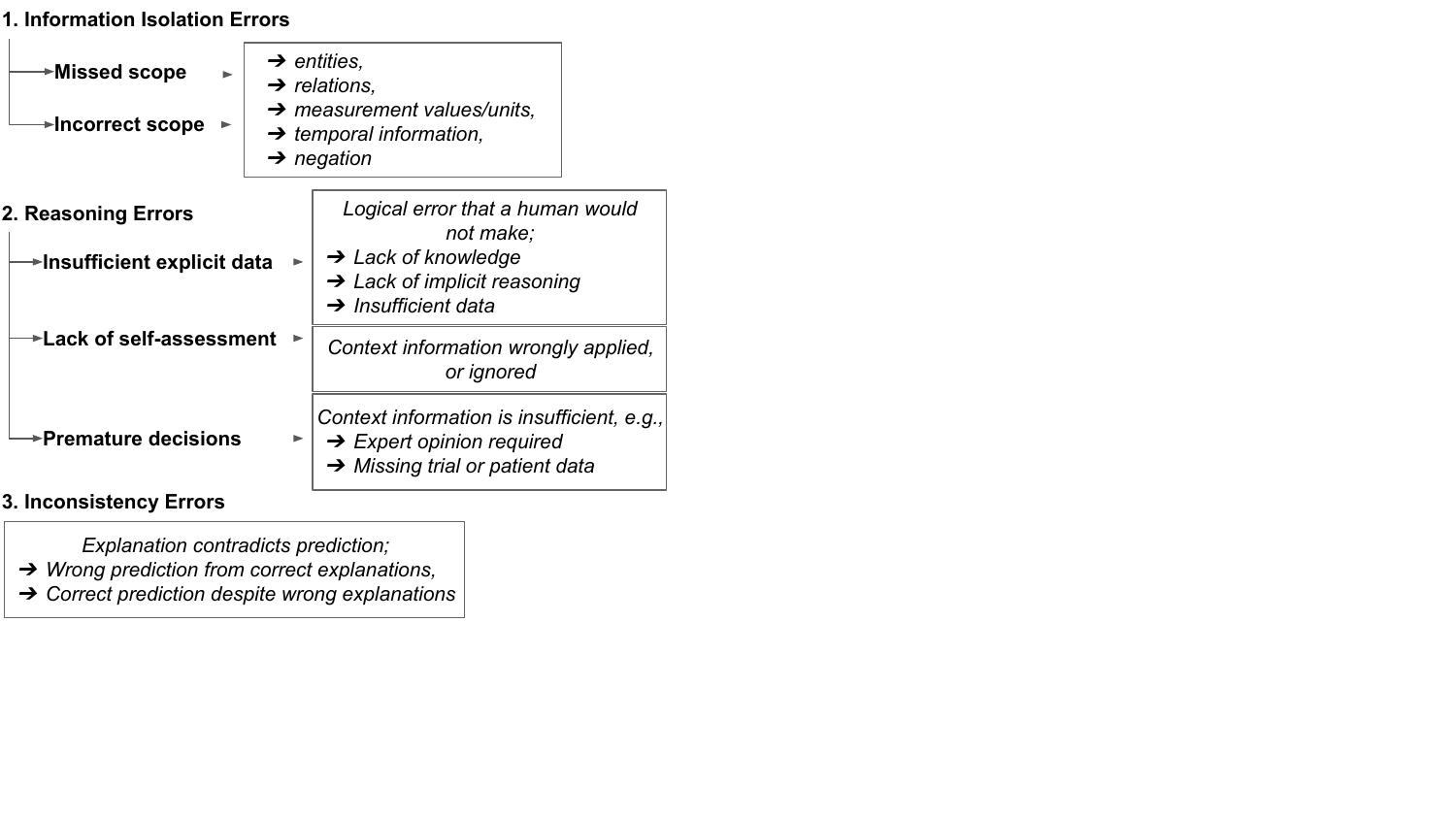}
    \caption{Taxonomy of errors in LLM-generations.}
    \label{fig:taxonomy-of-errors}
\end{figure}

\subsection{Taxonomy of LLM Errors}\label{sec:taxonomy_of_errors}
Since there is no one accepted taxonomy of errors, we often come across inconsistent and semantically overlapping categories of errors in LLM generated explanations. 
For e.g., \textit{incorrect reasoning} and \textit{lack of knowledge} are recognized as independent error types by \citet{jin2024matching, hamer2023improving, nievas2024distilling}, even though the latter often leads to the former (full list in Appendix \ref{appen:error_dist}). 
In the proposal by \citet{lievin2024can}, reasoning errors are separate from reading comprehension, even though the instances of incorrect reading comprehension occur when the model ignores contextual information and \textit{``reasons''} using its learned knowledge. 
Here, we identify where errors occur and break them down further by an identifiable source of the error (see Figure \ref{fig:taxonomy-of-errors}). 

\paragraph{Information Isolation Errors} These are errors in information extraction from patient data or trial criteria falls. This includes \textbf{\textit{missed}} or \textbf{\textit{incorrect}} NER, measurements (numeric or unit), temporal scopes and negations. LLMs are good at recognizing entities and measurements, but still struggle with negations \cite{nievas2024distilling}. 

\paragraph{Reasoning Errors} An error in the explanations provided by LLMs falls into this category. The most common source is \textbf{\textit{insufficient explicit data}}, which occurs when 
LLMs fail to draw logical conclusions from given data, while an expert can. This stems from {previously unseen data} (\textit{``lack of knowledge''}) or the {inability to recall} prior information (\textit{``lack of implicit reasoning''}) or the inability to infer from context (\textit{``insufficient data''}).
The second source of reasoning errors is \textbf{\textit{lack of self-assessment}}. The LLM contradicts explicit information in the prompt. The error occurs when knowledge is wrongly recalled or knowledge is correctly recalled, but contextual information is not applied, resulting in wrong reasoning. This is often referred to as \textit{``incorrect knowledge''}.
The third source are \textbf{\textit{premature decisions}} made by an LLM when the criteria actually require information that is missing, such as, expert opinion.

\paragraph{Inconsistency Errors} 
Generating explanations to arrive at final answers unlock LLM's reasoning capacities \cite{wei2022chain}.
Even so, the explanation and the prediction can be inconsistent.
LLMs may predict incorrectly despite correct explanation (reported as explanation-output mismatch in \citet{nievas2024distilling}). The opposite situation, when a prediction is correct despite an incorrect explanation, is more difficult to evaluate. This shortcut, similar to cognitive biases in humans, hints towards a bias that the model picked up during training.
Both these cases negatively impact transparency and accountability of the matching system. 

\section{Discussion}\label{sec:discussion}
\paragraph{Annotated Corpora and their Size} 
Despite attempts to structure clinical trials \cite{chen2022knowledge}, finding similar trials and analyzing systematic failure cases is notoriously difficult \cite{rybinski2020clinical,white2023clinidigest}.
Criterion-level annotations require significant manual annotations, thereby limiting the size of the corpus. For instance, Chia \citet{kury2020chia} and LCT \cite{dobbins2022leaf}, are manually annotated corpora of 1000 trials each, while supervised annotators such as, EliIE \cite{kang2017eliie} and Criteria2Query \cite{yuan2019criteria2query}, require expensive manual labels (230 disease-specific clinical trials in this case).
Eligibility labels on real patient data from real enrollment status, necessary to discount reviewer bias, are only available on a small-scale, if at all, due to the trade-off between scale and privacy \cite{kim2022review}. Systems thus evaluated on private retrospective data \cite{wong2023scaling, yuan2023large, unlu2024retrieval} cannot be transparently compared.\citet{kantor2024machine} stress on standardized benchmarks as the dependence on manual evaluation hinders meta-analyses and comparison between different studies.
Public annotations, such as the TREC CT tracks have an average of 700 trial annotations per patient for less than 100 patients, while the N2C2 has only a few thousands of criterion-level annotations.
Jointly, the major public corpora, e.g., \textit{ClinicalTrials.gov} and \textit{MIMIC} datasets, present an opportunity to build on the proposal by \citet{kim2022review} to generate large-scale data using automated methods. 

\paragraph{Dimensions of Evaluation}
The benchmarks in Table \ref{tab:benchmarks_and_evaluation} focus on model accuracy, corroborating the result from \citet{bedi2025testing}, which reports that more than 95\% of studies use accuracy as the primary dimension of evaluation, while fairness, bias and uncertainty are less frequently measured.
\citet{omar2024large} reviewed 27 clinical trials evaluating LLMs in healthcare also found the accuracy and reliability standards for LLM use to be undefined.
Further results from \citet{nemati2025benchmarking}, who benchmarked the annotation ability of LLMs across 9 performance metrics, show that while LLMs consistently score high on precision, recall and F1 (lowest being 0.8), their scores highly vary on semantic similarity, factual consistency, relevance, fluency, consistency and coherence (ranging from 0.1 to 0.9) highlighting the need for multiple dimensions of evaluation.

\paragraph{Formalization}\label{subsec:discussion_formalization}
Equation \ref{eq:matching} highlights the importance of aggregation and priority.
Surprisingly, very few works explicitly formalize the matching task. This lack of formalization coincides with an absence of aggregation strategies for trial-level predictions \cite{hamer2023improving, wornow2025zero, unlu2024retrieval} and others. 
Formalization guides \citet{yuan2023large} to design a loss function that accounts for the contrastive requirements of inclusion and exclusion criteria, \citet{jullien2024controlled} and \citet{saeidi2025streamlining} to define re-ranking scores.



\paragraph{Societal Impact}
Recent research measured distinct bias in disease diagnosis across gender, age and disease in popular LLM models (GPT4, ChatGPT and Qwen) \cite{zhao2024can} and found that GPT4 tends to stereotype demographic presentations when generating diagnoses \cite{zack2024assessing}. \citet{alber2025medical} show that LLMs are prone to making medical misjudgments by replacing just 0.001\% of the training data with medical misinformation. 
All the LLM-based systems, except for the trial ranking models, are evaluated on a few disease-specific trials (see Table \ref{tab:systems_overview_llm} in Appendix \ref{appendix:table_matching_systems}), the largest being 146 trials, covering 10 cancer types, evaluated by \citet{hamer2023improving}, making the generalizability of LLMs across diseases unclear. 


\paragraph{Data Sensitivity}
While we found several methods that deployed GPT models on Azure AI to comply with privacy regulations \cite{unlu2024retrieval,wong2023scaling,wornow2025zero}, processing patient data with LLMs still raises serious ethical challenges due to the lack of HIPAA compliance \cite{edemekong2024health}. 
As already discussed, creating large-scale realistic patient records while protecting their privacy is particularly challenging.
There is interest in generating synthetic data, called digital twins, with limited access to real patient data as a viable privacy protected alternative \cite{das2023twin,wang2024twin}. 

\paragraph{Core Limitations of LLMs}
According to \citet{harrer2023attention} the core limitations affecting LLM adoption are \textbf{unfiltered pre-training}, which does not differentiate between facts, opinions, or misinformation; \textbf{lack of self-assessment}, where a model generates invalid but syntactically and semantically coherent sentences; \textbf{non-determinism}: where surface-form prompt variations lead to drastic changes in the output and repeatability is not guaranteed under consistent input conditions; and, \textbf{knowledge recall}: where updating outdated data or injecting new information requires expensive retraining since the mechanisms of memory in LLMs are not well understood. In Section \ref{sec:taxonomy_of_errors}, we discussed how knowledge recall and lack of self-assessment surface through reasoning errors. 
These limitations pose a direct challenge to the transparency and accountability principles of AI for health laid down by the World Health Organization \cite{guidance2021ethics}.

\section{Future Directions}\label{sec:future_directions}
As interest in LLMs orchestrating an end-to-end pipeline and incorporating human interactions is gaining more attention \cite{gao2024empowering, qiu2024llm}, we focus on four promising directions.



\paragraph{Trial Search}
Past trials that share a target population are important for designing new trials, recruiting patients, systematic reviews and meta-analyses. This problem has seen little activity since clustering using lexical features \cite{hao2014clustering}. Newer search methods include constructing a clinical trial knowledge graph \citet{chen2022knowledge}, searching via patient EHRs \cite{wu2018semehr} and designing features for similarity matching\citet{sun2022interactivess}. \citet{jullien2023nli4ct} use textual entailment in LLMs to find trials that match short descriptions. However, these models falter on inferences that require numerical reasoning and could not surpass a BM25 baseline for ranking evidence.

\paragraph{Interactive Trial Design}
LLM agents have the potential of bridging the semantic gap between eligibility criteria and patient data by suggesting data models underlying patient data for structuring eligibility criteria early on. 
This collaborative idea is not new \cite{luo2013human}, yet, designing eligibility criteria is a big challenge and has been handled post-hoc by optionally considering patient data models \cite{kang2017eliie, sun2019knowledge, liu2021knowledge, dasgupta2020extracting}.
Small patient pools, which ultimately affect the successful completion of a trial, are often the result of restrictive criteria \cite{clinicaltrialarena2022}. 
LLMs can improve trial design by identifying restrictive criteria for trial investigators to relax them and create larger patient pools, especially for trials tackling diseases with a high mortality rate \cite{liu2021evaluating}.

\paragraph{Collaborative Trial Planning}
\citet{liu2025autoct} propose an iterative feature discovery model using LLM agents for interpretable trial outcome prediction.
\citet{markey2025rags} showed promising results on content relevance and suitability of trial protocols generated using LLMs, with room for improvement in logical reasoning and provenance. Similar to \citet{shi2024enhancing}, who propose collaboration of agents for knowledge-augmentation and reasoning, LLM agents can identify and distribute tasks and aggregate them to a final result. Another important operation is learning to defer to experts \cite{mozannar2020consistent}, which can separate operable criteria, such as those that require tabular operations (e.g., via structured queries) or reasoning (temporal, numerical, negation), from criteria that require expert feedback. 

\paragraph{Explainable Matches} Explaining black-box LLM predictions in human-understandable form is very challenging \cite{zhao2024explainability}. Explainability in clinical trial matches is limited to chain-of-thought generations, which is only one of the many facets of explainability \cite{nauta2023anecdotal, bodria2023benchmarking, chen2024evaluating}. 
\citet{wong2023scaling} could potentially explain eligibility prediction via logical component (mis)matches.
Furthermore, the utility of these explanations are unclear, as they have been only evaluated manually. We hope that the error taxonomy discussed in Section \ref{sec:taxonomy_of_errors} assists in systematic evaluation of explanations.

\section{Conclusion}
The task of clinical trial recruitment, that matches patients to a clinical trial via its participation eligibility criteria, benefits from knowledge aggregation and reasoning abilities of LLMs. 
In this survey, we critically examine the evolving role of LLM technologies in clinical research. We analyze the main components in a clinical trial recruitment process and provide a modern perspective on the challenges in adopting LLMs to clinical research, such as the size of the benchmarks, the dimensions of evaluation and data sensitivity. 
We hope that this serves as a valuable resource for future research.

\section{Limitations}
This survey focuses on the role of advances in NLP in the critical domain of clinical trial recruitment. Given that this is a rapidly evolving field, we have made our best effort to include a comprehensive view of available resources and methods. 
It is possible that more sophisticated methods using the latest technology already exist (e.g., in the form of proprietary products), but are not yet made public or are only available as abstracts, as is common in some medical communities, for example, the Annual Meeting of the American Society of Clinical Oncology (ASCO). 

\bibliography{custom}


\appendix

\section{Systematic Reviews versus Surveys}\label{appen:survey_guidelines}
A systematic review is a common practice in the medical research community, with standardized reporting guidelines \cite{moher2009preferred, page2021prisma}. This guideline has a checklist of required items in the title, abstract, introduction, methods, results, discussion and other information that every study needs to fulfil. These studies are bound by strict inclusion and exclusion criteria applied to the scientific literature considered for any assessment. This rigor is necessary in evidence-based analysis of a specific research question. A survey is more flexible and provides a board coverage of the topic being discussed.
For instance, the guidelines for writing surveys in our community as outlined by 
\citet{ACMCSUR}, \citet{TACL} and \citet{ARR}
aim to draw perspectives on an evolving topic of interest.

\paragraph{ACM Computing Surveys (long paper)} 
A paper that summarizes and organizes recent research results in a novel way that integrates and adds understanding to work in the field. A survey article assumes a general knowledge of the area; it emphasizes the classification of the existing literature, developing a perspective on the area, and evaluating trends.

\paragraph{TACL (excerpt)} They should thus not simply be a descriptive enumeration of the contents of papers, but draw broad themes and (importantly) provide new insights on the topic. These insights should be major contributions of the submission.

\paragraph{ARR (note)} all papers are expected to include reviews of related literature. This category is meant for the papers that go beyond that, e.g. in scope or in establishing new interdisciplinary connections.

\section{Reported Error Types}\label{appen:error_dist}
In the following we illustrate the the error distributions by percentages as reported by previous studies.

\noindent Errors reported in \cite{jin2024matching} supplementary:
\begin{itemize}[noitemsep,labelindent=0pt,leftmargin=1em,topsep=0pt]
    \item Incorrect reasoning: 30.7\%
    \item Lack of medical knowledge: 15.4\%
    \item Ambiguous label definition: 26.9\%
    \item other errors: 26.9\% (self-conflicting)
\end{itemize}

\noindent Errors reported in \cite{hamer2023improving}:
\begin{itemize}[noitemsep,labelindent=0pt,leftmargin=1em,topsep=0pt]
    \item Incorrect reading: 6.5\%
    \item Insufficient knowledge: 2.2\%
    \item Incorrect reasoning: 91.3\%
\end{itemize}

\noindent Errors reported in \cite{nievas2024distilling}:
\begin{itemize}[noitemsep,labelindent=0pt,leftmargin=1em,topsep=0pt]
    \item Lack of knowledge ~55\%
    \item Implicit criteria ~15\%
    \item Wrong reasoning
    \item Accurate reasoning, wrong decision
    \item Lack of restraint when expert opinion is needed
    \item Negated criteria error
\end{itemize}


\section{Trial-Patient Matching Systems}\label{appendix:table_matching_systems}
Table \ref{tab:systems_overview_classical} gives an overview of all 
the classical systems in covered in this survey by their direction of approach, method, source data characteristics and limitations. Table \ref{tab:systems_overview_llm} covers all the LLM-based systems discussed in this survey by their task, method, data source characteristics and limitations. We notice the shift in the availability of patient sources towards public data in the LLM-based approaches. We also notice a shift towards trial ranking due to publicly available data.

\begin{table*}
    \begin{adjustbox}{width=\linewidth, center}
    \small
    \begin{tabular}{p{2.3cm} p{3cm} p{2.5cm} p{2.5cm} p{3cm}}
    \toprule 
    \textbf{Direction} & 
    \textbf{Method} &
    \textbf{Patient Source} & 
    \textbf{Trial Source} & 
    \textbf{Limitations} \\
    \midrule
    
    {Trial-centric} &
    \citet{penberthy2010automated} 
    Discrete-data filter and sub-word matching &
    \textbf{Availability}:\newline Private
    \newline \textbf{Size} 282-2,112
    \newline \textbf{Mode}: text, tables
    \newline \textbf{Additional annotation}: expert review &
    \textbf{Availability}:\newline public (1); unknown (4) 
    \newline\textbf{Size}: 5 
    \newline\textbf{Type}: Specific &
    Manually coded eligibility criteria.
    \newline String similarity for keyword search.\\

    \cmidrule{3-5}
    
    & \citet{yuan2019criteria2query} 
    (Criteria2Query) \newline NL to structured queries \newline
    Information extraction, normalization 
    \newline Mapping to pre-defined cohort templates &
    No patient data &
    \textbf{Availability}:\newline Public$^*$ 
    \newline\textbf{Size}: 10
    \newline\textbf{Type}: Varying
    \newline\textbf{Additional annotation}: 125 criteria sentence, 215 entities, 34 relations, 137 negations, 20 attributes &
    Disease-specific NER.
    \newline Query templates cannot deal with missing attributes.
    \newline Entity normalization into a small set of 2000 concepts.
    \\

    \cmidrule{3-5}
    &
    \citet{tun2023automatic}\newline
    Rule-based filters
    \newline Similarity scores per criteria as features to a classifier &
    \textbf{Availability}:\newline Private
    \newline\textbf{Size}: 40,000
    \newline\textbf{Mode}: text, tables
    \newline\textbf{Additional annotation}: 109 patients labels &
    \textbf{Availability}:\newline Unknown
    \newline\textbf{Size}: 1
    \newline\textbf{Type}: Specific &
    Limited to a single trial
    \newline Rules and classifiers are not robust to changes in criteria \\

    \cmidrule{3-5}
    &
    \citet{ni2015increasing}\newline
    Discrete-data filter
    \newline Index trial and patient bag-of-words vectors
    \newline Return top vector matches for a trial &
    \textbf{Availability}:\newline Private
    \newline\textbf{Size}: 215
    \newline\textbf{Mode}: text, tables 
    \newline \textbf{Additional annotation}: historical match, expert review &
    \textbf{Availability}:\newline Public$^*$
    \newline\textbf{Size}: 55
    \newline\textbf{Type}: Specific &
    12.6\% precision\\

    \cmidrule{2-5}
    Patient-centric
    &
    \citet{ni2015increasing}\newline
    Return top vector matches for a patient &
    \multicolumn{2}{c}{\textit{(same as previous row)}} &
    4\% precision
    \\
    
    \cmidrule{3-5}
    &
    \citet{zhang2017automated} \newline
    Bag-of-words feature vector
    \newline SVM classifier &
    No patient data
    &
    \textbf{Availability}:\newline Public$^*$
    \newline\textbf{Size}: 2461
    \newline\textbf{Type}: 891 Specific; 1570 Varying
    \newline\textbf{Additional annotation}: Trial labels  &
    Cohort-specific model
    \newline No real patient data considered
    \newline Closest to keyword search (trials to cohort)\\

    \bottomrule
    \multicolumn{5}{l}{\small$^*$ Public source for clinical trials: \url{https://clinicalTrials.gov}}
    \end{tabular}
    \end{adjustbox}
    \caption{Overview of classical systems covered in this survey: direction, methods, data sources and limitations. }
    \label{tab:systems_overview_classical}
\end{table*}

\onecolumn
{
\small
\begin{longtable}{@{}p{2cm} p{3cm} p{3cm} p{3cm} p{3cm}@{}}
    \toprule
    \textbf{Task} & 
    \textbf{Method} &
    \textbf{Patient Source} & 
    \textbf{Trial Source} & 
    \textbf{Limitations} \\
    \midrule    
    \endfirsthead
    
    \toprule 
    \textbf{Task} & 
    \textbf{Method} &
    \textbf{Patient Source} & 
    \textbf{Trial Source} & 
    \textbf{Limitations} \\
    \midrule  
    \endhead

    \midrule
    \multicolumn{5}{l}{\textit{Continues to the next page.}}\\
    \bottomrule
    \endfoot

    \bottomrule
    \multicolumn{5}{l}{\textit{$^*$ EudraCT (European Union Drug Regulating Authorities Clinical Trials) is the European clinical trials database \cite{eudract}.}}\\
    \multicolumn{5}{l}{\textit{$^{**}$ Public source for clinical trials: \url{clinicalTrials.gov}.}}\\
    \caption{Overview of LLM-based systems covered in this survey: tasks, methods, data sources used and limitations.}
    \label{tab:systems_overview_llm}
    \endlastfoot

    Criterion-level prediction &
    \citet{hamer2023improving}\newline 1-shot prompt to LLM for criteria level prediction with explanation &
    \textbf{Availability}: Private (synthetic patient profile)
    \newline\textbf{Size}: 10
    \newline\textbf{Mode}: Short text
    \newline\textbf{Type}: Specific &
    \textbf{Availability}: Mixed (clinicalTrials.gov, EudraCT$^*$)
    \newline\textbf{Size}: 146 clinical trials
    \newline\textbf{Type}: Specific
    \newline\textbf{Additional Annotation}: Expert review &
    Majority of criterion-level errors due to incorrect reasoning (91\%)\\
    
    \cmidrule{3-5}
    &
    \citet{unlu2024retrieval}\newline RAG-based &
    \textbf{Availability}: Private
    \newline\textbf{Size}: 2,276 
    \newline\textbf{Mode}: Tables; text &
    \textbf{Availability}: Public$^{**}$ 
    \newline\textbf{Size}: 1 
    \newline\textbf{Type}: Specific &
    RAG pipeline not published. Evaluated on a single trial.\\

    \cmidrule{3-5}
    &
    \citet{ferber2024end}
    \newline Criterion-level boolean eligibility labels &
    \textbf{Availability}: Private
    \newline\textbf{Size}: 51 
    \newline\textbf{Mode}: EHR &
    \textbf{Availability}: Public$^{**}$ 
    \newline\textbf{Size}: 105,600
    \newline\textbf{Type}: Specific &
    Use of proprietary model limits evaluation on synthetic patient data.\\

    \cmidrule{3-5}
    &
    \citet{beattie2024utilizing}
    \newline Vector database of patient notes. Prompt (criteria, top-k patient notes) pair to LLM &
    \multicolumn{2}{l}{\textbf{Availability}: Public (2018 N2C2: Cohort Selection)} &
    Some evaluation is on test subset (40 of 182).
    \newline Expert guidance requires manual expertise for every criterion.\\ 
    
    \cmidrule{3-5}
    &
    \citet{wornow2025zero}\newline Vector database of patient notes. Prompt (criteria, top-k patient notes) pair to LLM &
    \multicolumn{2}{l}{\textbf{Availability}: Public (2018 N2C2: Cohort Selection)}&
    Doesn't consider exclusion criteria
    \newline 67\% incorrect decision despite correct reasoning.\\

    \cmidrule{3-5}
    &
    \citet{saeidi2025streamlining}
    \newline Prompt LLM with a fine-tuned BERT-based concept embeddings from patient and criteria to predict binary eligibility labels. &
    \multicolumn{2}{l}{\textbf{Availability}: Public (2018 N2C2: Cohort Selection)} &
    LLM reasoning capabilities not explored.\\
    
    \cmidrule{2-5}    
    Trial-level prediction &
    \citet{wong2023scaling}\newline LLM prompted with a template to generate structured forms of eligibility criteria &
    \textbf{Availability}: Private
    \newline\textbf{Size}: Unknown
    \newline\textbf{Mode}: Structured
    \newline\textbf{Additional Annotation}:
    523 trial-patient historical labels;
    68,485 trial-patient new labels&
    \textbf{Availability}: Public$^{**}$ 
    \newline\textbf{Size}: 53
    \newline\textbf{Type}: Specific &
    Clinical Trial eligibility limited to the first 40 lines.
    \newline Trial-level performance is very low (F1 score range:[29.6-48])\\

    \cmidrule{3-5}
    &
    \citet{yuan2023large} 
    Prompt LLMs to reformulate eligibility criteria &
    \textbf{Availability}: Private
    \newline\textbf{Size}: 825 
    \newline\textbf{Mode}: Longitudinal text records&
    \textbf{Availability}: Public$^{**}$
    \newline\textbf{Size}: 6
    \newline\textbf{Type}: Specific
    \newline\textbf{Additional Annotation}: 100,000 criterion-patient labels &
    Criterion-level performance metrics 10 points higher than trial-patient.
    \newline High variance between different trials: F1 range: 0.39 - 0.96.\\

    
    \cmidrule{2-5}
    Trial ranking &

    \citet{pradeep2022neural}\newline Synthesis queries for initial trial retrieval. \newline Fine-tune T5 to generate relevance label based on patient description and trial data &
    \multicolumn{2}{l}{\textbf{Availability}: Public (TREC CT 2021)} &
    Zero-shot relevance ranking is only slightly better than BM25. Fine-tuned models need re-training on new data or format.\\

    \cmidrule{3-5}
    &
    \citet{zhuang2023team}\newline Hybrid sparse-dense retriever for top-1000.
    \newline Cross-encoder re-ranker
    \newline GPT4 for top-20 re-rank &
    \multicolumn{2}{l}{\textbf{Availability}: Public (TREC CT 2023)} &
    High NDCG, 0.51 precision, 0.38 recall (very low)
    \newline Validation on textual patient note does not transfer well to testing on structured patient note\\

    \cmidrule{3-5}
    &
    Cosine similarity between patient and trial embeddings obtained using GPT by \citet{richmond2023leveraging} and using Sentence Transformer by \citet{lahiri2023tmu}.
    & \multicolumn{2}{l}{\textbf{Availability}: Public (TREC CT 2023)} &
    Very low MAP of 0.02 \cite{richmond2023leveraging} and close to zero for \citet{lahiri2023tmu} \\

    \cmidrule{3-5}
    &
    \citet{kusa2023dossier}\newline Sentence query formulation using GPT-3.5. Query enrichment using \citet{kusa2023effective}. Trial-level prediction of re-ranked pairs using GPT-3.5.
    & \multicolumn{2}{l}{\textbf{Availability}: Public (TREC CT 2023)} &
    Zero-shot LLM prompts provide marginal improvement over neural-rerankers\\

    \cmidrule{3-5}
    &
    \citet{piekos2023unimib} \newline Utilize GPT3.5 to obtain trial-level labels for final re-ranking on top of lexical and neural re-rankers & 
    \multicolumn{2}{l}{\textbf{Availability}: Public (TREC CT 2023)} &
    Query processing is template-based and excludes negative information. \\
    
    \cmidrule{3-5}
    &
    \citet{ferber2024end}
    \newline No-SQL query formulation from patient EHR using GPT-4o for initial retrieval, followed by vector embedding re-ranking. &
    \textbf{Availability}: Private
    \newline\textbf{Size}: 51 
    \newline\textbf{Mode}: EHR &
    \textbf{Availability}: Public$^{**}$ 
    \newline\textbf{Size}: 105,600
    \newline\textbf{Type}: Specific &
    Use of proprietary model limits evaluation on synthetic patient data.\\

    \cmidrule{3-5}
    &
    \citet{rybinski2023matching,rybinski2024learning}\newline Multi-stage retriever using LLMs &
    \multicolumn{2}{l}{\textbf{Availability}: Public (TREC CT 2021, 2022, 2023)} &
    High latency of GPT4 vs a small performance boost\\

    \cmidrule{3-5}
    &
    \citet{jullien2024controlled}
    \newline LLM-guided basic attribute extraction for trial retrieval and filtering. Criterion-level LLM predictions fed to set-reasoning-based re-rank scoring functions. 
    &
    \multicolumn{2}{l}{\textbf{Availability}: Public (TREC CT 2022)} &
    LLMs underperform in exclusion criteria labelling.\\
    
    \cmidrule{3-5}
    &
    \citet{jin2024matching}
    \newline(TrialGPT)
    \newline Hybrid trial filtering with BM25 and MedCPT \cite{jin2023medcpt}
    \newline Criterion-level LLM  prediction 
    aggregated to trial scores  &
    \multicolumn{2}{p{6cm}}{\textbf{Availability}: Public (\citet{koopman2016test}, TREC CT 2021, 2022)
    \newline \textbf{Additional annotation}: 1,015 criterion-patient labels }
    &
    User study sample size is small
    \newline Does not handle longitudinal patient data
    \newline LLM aggregation requires LLM to perform mathematical reasoning and computations\\

    \cmidrule{3-5}
    &
    \citet{nievas2024distilling}
    \newline Extends \cite{jin2024matching} to open-source LLMs &
    \multicolumn{2}{p{6cm}}{\textit{(same as \citet{jin2024matching})}
    \newline\begin{tabular}{@{}p{3cm}p{3cm}@{}}
        \textbf{Additional annotation}: Patient sentence supporting eligibility label  &  \textbf{Additional annotation}: 500 criteria labels on: eligibility and difficulty.\\
    \end{tabular}}
    &
    Significantly high fine-tuning costs
    \\

    \cmidrule{3-5}
    &
    \citet{datta2025patient2trial}
    \newline (Patient2Trial)
    \newline Lexical retrievers use LLM generated query expansion. LLM predicts trial-level label and a criterion-level rationale with a matching score. Final ranking based on matching score. 
    \newline &
    \multicolumn{2}{l}{\textbf{Availability}: Public (TREC CT 2023)} &
    Trials prefiltered by disorder-specific keywords.
    \newline Manually curated disorder-specific instructions.
    \newline Model predicted trial-level label is not evaluated\\

    \cmidrule{3-5}
    &
    \citet{saeidi2023malnis,saeidi2025streamlining}
    Embed patients and trials to concept vector space. Use variations of Equation \ref{eq:matching} to compute relevance scores.
    &
    \multicolumn{2}{l}{\textbf{Availability}: Public (TREC CT 2023)} &
    No direct connection between criterion-level LLM predictions and trial-level relevance score computations\\

\end{longtable}
}

\twocolumn

\end{document}